\documentclass[format=acmsmall, review=false, screen=true]{acmart}

\usepackage{booktabs} 

\newcommand{\etal}{\mbox{\emph{et al.\ }}}
\newcommand{\ie}{\mbox{\emph{i.e.}}}
\newcommand{\eg}{\mbox{\emph{e.g.}}}

\usepackage[ruled]{algorithm2e} 

\usepackage{amsmath}
\usepackage{multirow}
\usepackage{color}
\usepackage{hyperref}
\usepackage{graphicx}
\usepackage{subcaption}

\usepackage{pifont}
\newcommand{\cmark}{\ding{51}}%

\SetAlFnt{\small}
\SetAlCapFnt{\small}
\SetAlCapNameFnt{\small}
\SetAlCapHSkip{0pt}
\IncMargin{-\parindent}

\newlength\savewidth

\acmJournal{TOMM}
\acmVolume{5}
\acmNumber{8}
\acmArticle{39}
\acmYear{2023}
\acmMonth{10}
\copyrightyear{2009}

\setcopyright{acmlicensed}

\acmDOI{10.1145/3656046}

\received{October 2023}
\received[revised]{xx xxxx}
\received[accepted]{xx xxxx}
\begin{document}
\author{Xiaolong Shen}
\affiliation{%
  \institution{Zhejiang University}
  \postcode{310013}
  \country{Zhejiang, China}
  }
  
\author{Zhedong Zheng}
\affiliation{%
  \institution{University of Macau}
  \postcode{999078}
  \country{Macau SAR, China}
  }
\author{Yi Yang}
\affiliation{%
  \institution{Zhejiang University}
  \postcode{310013}
  \country{Zhejiang, China}
  }

\title{StepNet: Spatial-temporal Part-aware Network for Isolated Sign Language Recognition}

\markboth{Journal of \LaTeX\ Class Files,~Vol.~14, No.~8, August~2015}%
{Shell \MakeLowercase{\textit{et al.}}: Bare Demo of IEEEtran.cls for Computer Society Journals}

\newcommand{\zznote}[1]{\textcolor{black}{#1}}

\begin{abstract}
The goal of sign language recognition (SLR) is to help those who are hard of hearing or deaf overcome the communication barrier. Most existing approaches can be typically divided into two lines, \ie, Skeleton-based and RGB-based methods, but both the two lines of methods have their limitations. Skeleton-based methods do not consider facial expressions, while RGB-based approaches usually ignore the fine-grained hand structure. To overcome both limitations, we propose a new framework called Spatial-temporal Part-aware network~(StepNet), based on RGB parts. As its name suggests, it is made up of two modules: Part-level Spatial Modeling and Part-level Temporal Modeling. Part-level Spatial Modeling, in particular, automatically captures the appearance-based properties, such as hands and faces, in the feature space without the use of any keypoint-level annotations. On the other hand, Part-level Temporal Modeling implicitly mines the long-short term context to capture the relevant attributes over time. Extensive experiments demonstrate that our StepNet, thanks to spatial-temporal modules, achieves competitive Top-1 Per-instance accuracy on three commonly-used SLR benchmarks, \ie, 56.89\% on WLASL, 77.2\% on NMFs-CSL, and 77.1\% on BOBSL. Additionally, the proposed method is compatible with the optical flow input and can produce superior performance if fused. For those who are hard of hearing, we hope that our work can act as a preliminary step.
\end{abstract}



\thanks{
Xiaolong Shen and Yi Yang are with the College of Computer Science and Technology, Zhejiang
University, China 310027. E-mail: sxlongcs@zju.edu.com, yangyics@zju.edu.cn.
Zhedong Zheng is with Faculty of Science and Technology, and Institute of Collaborative Innovation, University of Macau, China 999078. E-mail: zhedongzheng@um.edu.mo.

This work was supported by the National Natural Science Foundation of China (U2336212) and the Fundamental Research Funds for the Central Universities (No. 226-2022-00051).
}

\maketitle
\renewcommand{\shortauthors}{X. Shen et al.}

\section{Introduction}
	
	As the primary communication tool for deaf and hard of hearing, sign language employs the dynamic movement of the hands and body as well as facial expressions to convey meaning.
	Sign language has complex rules and is challenging to learn because it is independent of spoken language.
	Moreover, sign languages are not universal although there are some similarities among different sign languages~\cite{wiki-sign-language}. 
	It makes a communication barrier between deaf-mute people and others, as well as among deaf-mute people.
	In an attempt to break such barriers, more researchers~\cite{li2020word,hu2021signbert,jiang2021sign} have started to explore Sign language recognition (SLR), which aims to predict the word class of the sign video. SLR is of importance, since it is the fundamental prerequisite of sign language translation~\cite{guo2019hierarchical,camgoz2020sign} and other related tasks, such as sign language generation~\cite{suo}.  
	
	\begin{figure}[t]
		\centering
		\includegraphics[width=1\linewidth]{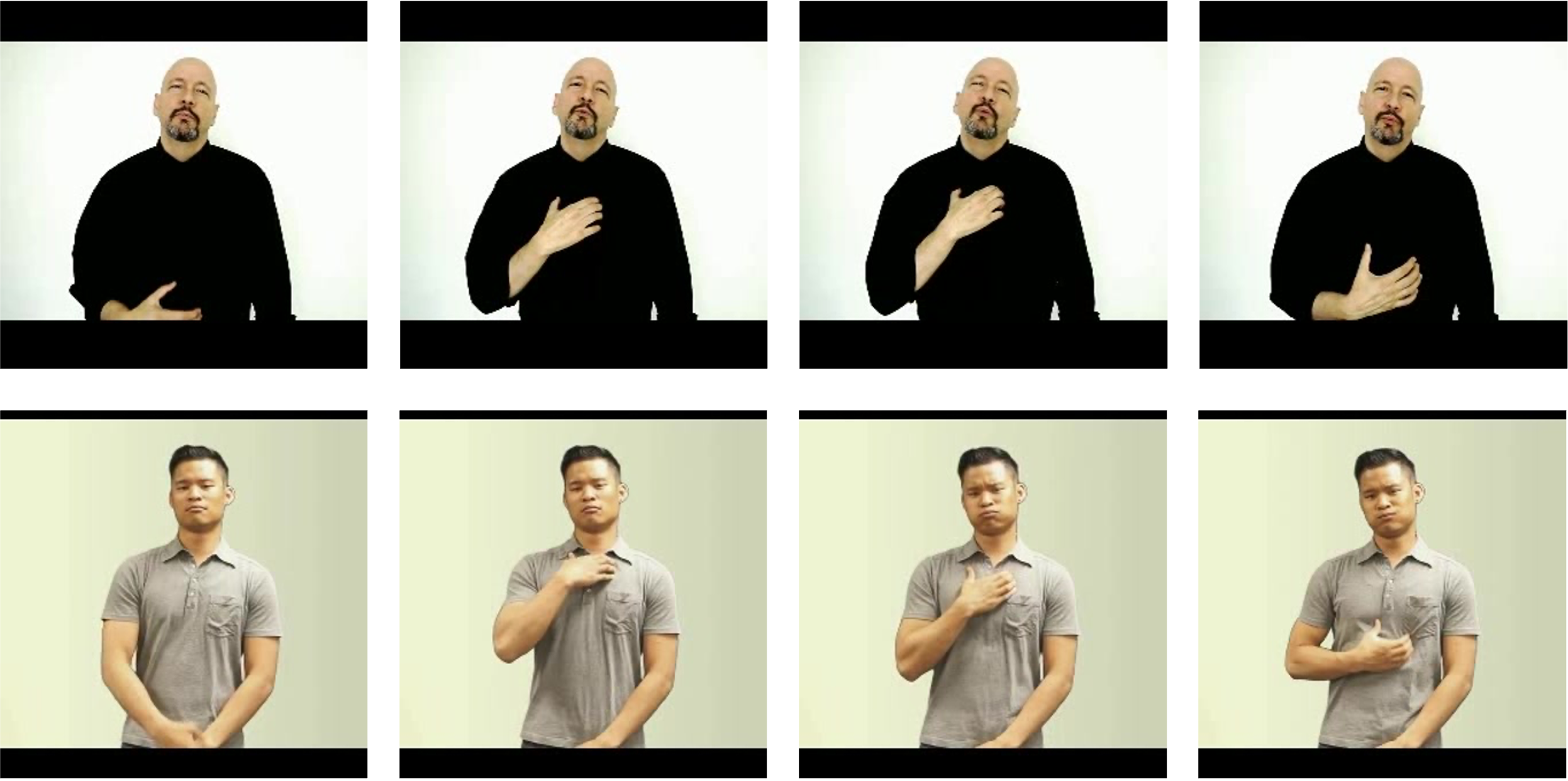} 
		\caption{Selected challenging examples from the widely-used sign language WLASL benchmark~\cite{li2020word}. We could observe that there are some similar-appearance hand gestures, \eg, ``wish"~(top) and ``hungry"~(bottom). Such cases demand the learned model to mine more fine-grained details, such as facial expressions. 
		}
		\label{same_sign}
	\end{figure}
	
	Therefore, in this work, we study word-level sign language recognition, called Isolated SLR. Isolated SLR is generally regarded as a fine-grained action recognition task.
	Previously, SLR approaches~\cite{grobel1997isolated,wang2014similarity,koller2015continuous,tharwat2015sift} encode hand gestures with hand-crafted features. 
	Recently, owing to the development of deep learning, many methods~\cite{jiang2021sign,hu2021signbert,hu2021global} have emerged for learning a sign video representation in an end-to-end manner. 
	These works generally involve two families based on the input format: Skeleton-based (using the keypoint sequence) and RGB-based (using the RGB video) methods.
	Skeleton-based methods~\cite{pigou2018beyond,sincan2019isolated,li2020word,jiang2021sign} mainly utilize Recurrent neural networks (RNNs) or Graph neural networks (GNNs) to model the spatial and temporal representation. 
	However, the Skeleton-based approaches have two primary drawbacks. First, appearance attributes are totally discarded in the keypoint inputs, which is sub-optimal. 
	The facial expressions in SLR convey the emotion of the signer. The emotion can serve as an additional clue to discriminate the sign words with similar hand or limb movements (see Figure~\ref{same_sign}). 
	Second, keypoint annotations are not stable and usually noisy. For instance, the keypoints~\cite{li2020word,jiang2021sign} are extracted by an offline pose estimator. 
	Due to the gap between different training and testing sets, the offline keypoint model could introduce incorrect predictions, especially in the video scenario, \eg, the occlusion on the fingers and the motion blur. 
	In contrast, another line of works, \ie, RGB-based methods~\cite{li2020transfer,li2020word,albanie2020bsl}, mainly fine-tune the action recognition methods. Some works~\cite{li2020word}  deploy 2D Convolutional neural networks (CNNs) with RNNs to build spatial and temporal relations respectively, while others~\cite{li2020word,albanie2020bsl} also apply 3D CNNs to capture spatial as well as temporal information simultaneously. 
	However, directly borrowing the structure of the general action recognition framework ignores the geometric characteristics of sign language videos, such as fine-grained movement and human structure. 
	As shown in Figure~\ref{same_sign}, most pixels in sign videos are static, while the discriminative parts only take a few spaces in the frame. Therefore, the non-trivial slight hand movement and facial expression need ad-hoc optimization. 
	\begin{figure}[t]
		\centering
		\includegraphics[width=1\linewidth]{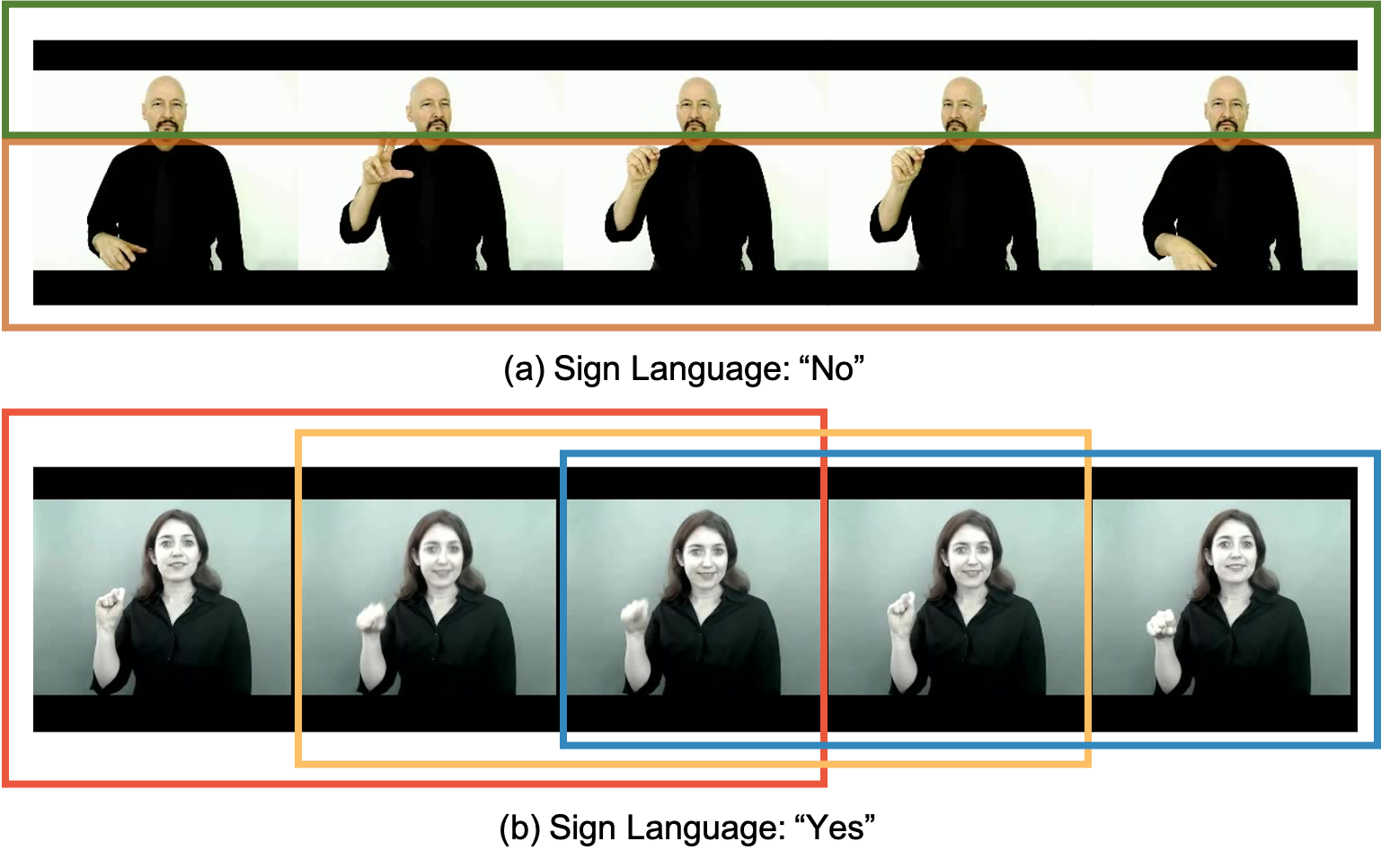} 
		\caption{Our motivation for designing the part-aware modules. (a) Part-level Spatial Partition: We can infer what the sign video means by looking at just one part, either the \textcolor{green}{green} part (shaking head) or the \textcolor{brown}{brown} part (closing three fingers) 
			(b) Part-level Temporal Partition: Similarly, we observe that any three short video clips, \ie, \textcolor{red}{red}, \textcolor{yellow}{yellow}, and \textcolor{blue}{blue}, can represent the gloss of the sign video. 
			These two kinds of partition inspire us to harness RGB parts to mine fine-grained features for sign language recognition.
		}
		\label{mo}
	\end{figure}
	
	Based on the above observation, we propose a new SLR network, called Spatial-temporal Part-aware network (StepNet), which is based on RGB parts instead of commonly-used skeleton data. It is because RGB parts well preserve the local human structure movement as well as the fine-grained texture changes, which is in line with the demands of the SLR task. As the name implies, StepNet contains two different part modeling branches. One branch is Part-level Spatial Modeling, and the other is Part-level Temporal Modeling. Part-level Spatial Modeling is to capture the appearance-based properties, such as hands and faces, in the feature space, while Part-level Temporal Modeling aims at capturing the changes along with the time (see Figure~\ref{mo}).
	When inference, we fuse the spatial and temporal representations for the final prediction. Since we involve two kinds of fine-grained branches, the learned model is robust to similar classes with subtle differences and achieves competitive performance on several public benchmarks~\cite{li2020word,albanie2021bbc,hu2021global}. 
	The ablation studies also show that our model effectively exploits the Spatial-temporal cues in sign videos. 
	Our main contributions are as follows: 
	\begin{itemize}
		
		\item Inspired by the inherent structure of sign language, we design a new RGB-based network, called Spatial-temporal Part-aware network (StepNet). As the name implies, StepNet contains the Part-level Spatial Modeling that learns the symmetric and top-bottom association, and the Part-level Temporal Modeling that captures bottom-up long-short term changes.  
		\item Extensive experiments verify the effectiveness of the proposed method, yielding competitive performance on public benchmarks, \ie, 77.1\% on BOBSL~\cite{albanie2021bbc}, 56.89\% on WLASL-2000~\cite{li2020word} and 77.2\% on NMFs-CSL~\cite{hu2021global}. Moreover, StepNet is compatible with the optical flow inputs for the further ensemble.
	
	\end{itemize}
	
	
\section{Related work} 
	\noindent\textbf{Sign language recognition (SLR).}
	Depending on the input format, SLR can be classified into RGB-based and Skeleton-based approaches. 
	Previously, pioneering RGB-based methods~\cite{buehler2009learning,cooper2012sign,liwicki2009automatic,yasir2015sift} mainly use hand-crafted features, \eg, HOG-based features, and SIFT-based features, to model spatio-temporal representation.
	Recently, deep learning has shown great potential in various computer vision tasks. Some approaches~\cite{koller2016deep,koller2018deep} utilize 2D CNNs for spatial feature extraction instead of hand-crafted features and Hidden Markov Models (HMMs) for temporal modeling. 
	In addition,~\cite{li2020word,albanie2020bsl,hosain2021hand} deploy 3D CNNs for spatio-temporal modeling, which achieves higher recognition accuracy. It also reflects the powerful characterization capability of 3D CNNs.
	Skeleton-based methods~\cite{bohavcek2022sign,tunga2021pose,li2020word} have attracted much interest from researchers thanks to the fact that skeleton data is not affected by background and appearance.
	Similar to the method of processing RGB data, some methods~\cite{du2015hierarchical,li2018co,cao2018skeleton,song2017end} utilize CNNs or RNNs to process skeleton data. Besides, there are other processing methods for skeleton data. For instance, one line of works~\cite{jiang2021sign,bohavcek2022sign,li2020word} utilizes GNNs to process skeleton data. 
	Taking one step further, SAM~\cite{jiang2021sign} employs the multi-stream strategy~\cite{shi2020skeleton} that separates the skeleton data into joint, bone, joint motion, and bone motion, and then applies the designed SL-GCN to model these representations. Recently, the pretraining and finetuning schema has been widely used in computer vision tasks. For sign language recognition, there are some works\cite{hu2021signbert, selvaraj2021openhands, Hu_2023, zhao2023best} that focus on this schema. Owing to this self-reconstruction schema in the pretraining stage, the model will mine the inherent relations as much as possible. Pretraining a model on a large dataset will learn a more powerful representation, improving the robustness of the model, particularly in some challenging cases like self-occlusion among joints and motion blur. Thus when finetuning this model on other downstream tasks, the model can leverage more dense information to make the choice. However, the Skeleton-based approaches have two primary drawbacks. First, appearance attributes are totally discarded in the keypoints input. Second, keypoint annotations are not stable and usually noisy. As for the existing RGB-based methods, they do not fully exploit the relation between face and hand, except that borrowing the structure of the general action recognition framework.
	\begin{figure*}[!ht]
		\centering
		\includegraphics[width=1.0\textwidth]{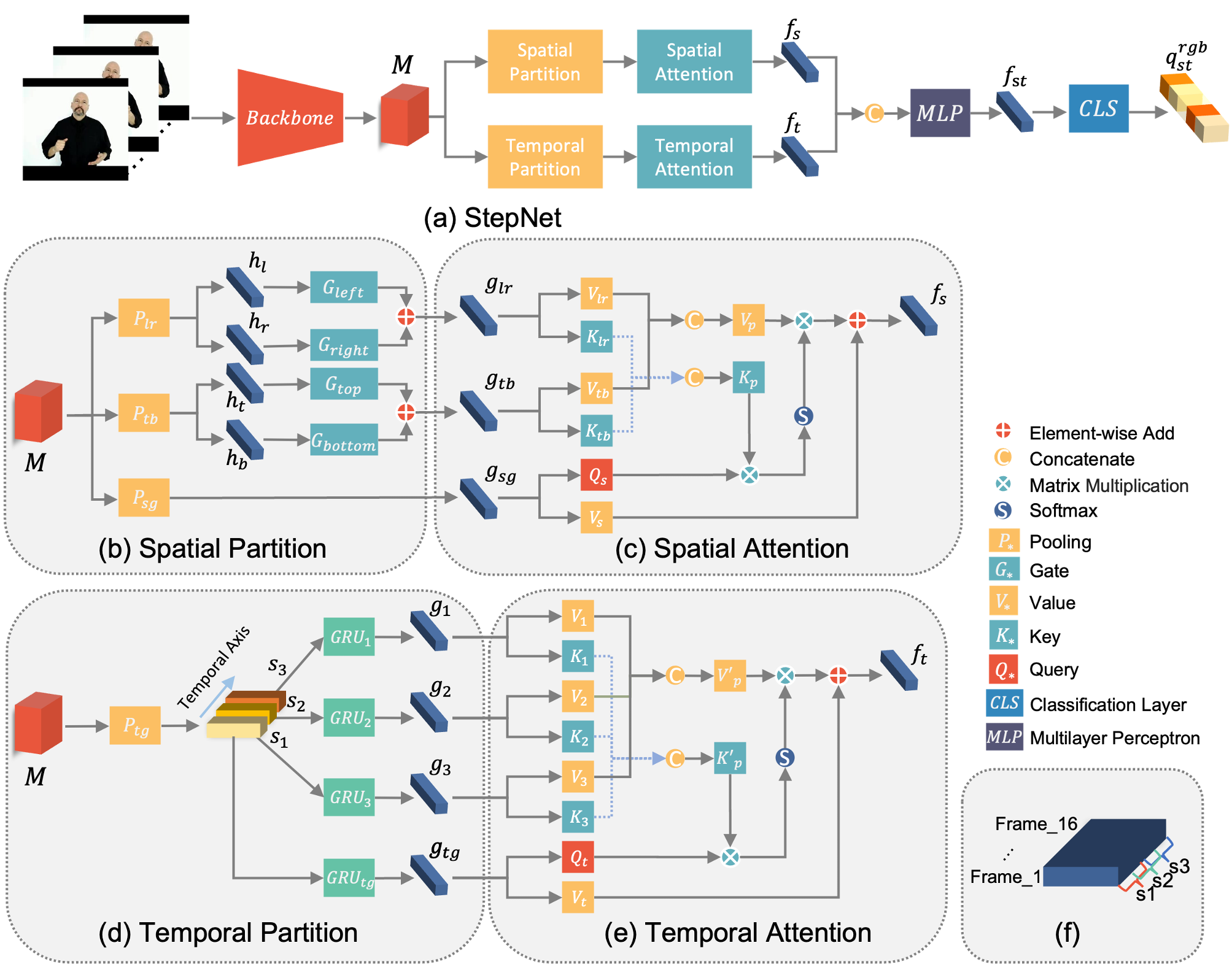}
		\caption{(a) A schematic of our framework. StepNet consists of four parts, \ie, backbone, Part-level Spatial Modeling, Part-level Temporal Modeling, and classification layer.
    The backbone is utilized to extract feature maps of inputs $\textbf{M}$. The feature maps $\textbf{M}$ are then processed in parallel using a Part-level Spatial Modeling and a Part-level Temporal Modeling. Finally, outputs of two part-level modeling are fused and put into the classification layer to obtain classification logit vectors~$q^{rgb}_{st}$.
			Sub-figure (b),(c). Details of the spatial partition and attention. Spatial partition includes the local guidance, \ie, left-right(lr) and top-bottom (tb), and the global guidance (sg). The lr and tb are composed of Pooling~(P) and Gate~(G) operators, which build the channel relationships between the lr or tb. Global guidance(sg) guides the model to obtain a coarse but global representation $g_{sg}$.
			After that, we introduce spatial attention in Sub-figure~(c) to compute how to aggregate these features $g_{lr}, g_{tb}, g_{sg}$.
			Sub-figure (d),(e). Details of the temporal partition and attention. We first pool the feature maps and then split them along the temporal dimension into three segments, \ie, $s_1, s_2, s_3$. Each segment is utilized to explore the short clip context by $GRU$. In addition, we also apply the $GRU$ to the long clip (before partition), which models the long-term representation~$g_{tg}$. After that, temporal attention learns how to complement the long-term representation~$g_{tg}$ from the short-term context. Sub-figure (f). An example of partition with overlaps along the temporal axis.
		}
		\label{framework}
	\end{figure*}
	
	\noindent\textbf{RGB-based action recognition.} The methods in this task mainly deploy three types of neural networks: 3D CNNs, 2D CNNs, and the mixtures of 3D and 2D.
	~\cite{carreira2017quo} proposes I3D to learn spatio-temporal features from videos directly. TEINet~\cite{liu2020teinet} and TEA~\cite{li2020tea} design more powerful temporal modules that can be directly incorporated into 2D CNNs for efficient information interaction between neighboring frames.   
	For example, TSM~\cite{lin2019tsm} shifts the channels along the temporal dimension, which enhances the ability of temporal modeling for 2D CNNs. R(2+1)D~\cite{tran2018closer} and S3D~\cite{xie2018rethinking} decompose the 3D convolution into a 2D spatial convolution and 1D temporal convolution to improve the efficiency of 3D convolution.
	Two-path networks are also effective 3D designs. For instance, SlowFast~\cite{feichtenhofer2019slowfast} applies a slow branch and a fast branch to capture long-short term temporal information by feeding video with different frame rates.
	
	
	\noindent\textbf{Multi-cues methods.} 
	To improve the effectiveness of models for sign language recognition, researchers explore various fusion methods~\cite{gammulle2021tmmf,guo2017online} that utilize the combination of diverse cues, including multiple modalities and multiple local patterns. 
	As one of the pioneering works, the two-stream method~\cite{simonyan2014two} uses two kinds of inputs, \ie, RGB and optical flow, to model appearance and motion information in videos separately and then fuse them with the late-fusion strategy. Similarly,
	SAM~\cite{jiang2021sign} designs a multi-branch framework that fuses RGB and depth-based modalities by the proposed GEM, yielding competitive performance. For sign language translation, ~\cite{guo2019hierarchical} proposes a temporal encoder to capture RGB and skeleton clues adaptively. Another line of work leverages local patterns within the input to mine the fine-grained feature.
	Hu~\etal~propose a model~\cite{hu2021hand} that refines the representation of the hand by the hand prior, which enhances the model interpretability, and then fuses the hand representation with RGB-based or Skeleton-based methods by late fusion to improve the model accuracy. Zhou~\etal~propose a spatial-temporal multi-cue network~\cite{zhou2021spatial} for continuous SLR and sign language translation which involves multiple customized modules and optimization strategies for multi-cue fusion. 
	Moreover, some methods~\cite{zhao2021learning,yang2020sta,9895208, 9288873,duan2018unified, liu2021imigue, fan2022pstnet,ding2021rfnet,yang2021collaborative,quan2024psychometry} in other fine-grained tasks also adopt a similar multi-cues strategy for learning discriminative representation. For instance, PCB~\cite{sun2018beyond} horizontally splits the feature maps,  while LPN~\cite{wang2021LPN} cuts the local attention parts in a circle format. Yu \etal propose a method~\cite{Yu_2021} that searches multi-rate and multi-modal architectures for RGB-D gesture recognition, and designs a 3D-CDC block to capture temporal context. Liu \etal propose a video dataset~\cite{liu2021imigue} for micro-gesture understanding and emotion analysis and design an unsupervised framework for MG Recognition.
 Some works~\cite{zheng2018pedestrian,guan2020thorax} introduce spatial transform on feature maps. In this work, we follow a similar multi-cue spirit. Differently, we focus on the symmetric and top-bottom association, and learning temporal modeling in an end-to-end manner. Therefore, the proposed method achieves competitive performance. For sign language recognition, using external tools to detect faces or hands and then feeding into the model separately to learn relations incur expensive computational overhead for Sign Language tasks.
	
\section{Methodology} \label{Methodology}
	
	\subsection{Overview}
	As shown in Figure~\ref{framework}, StepNet consists of three stages: the first stage is to extract the coarse spatiotemporal representation from the input RGB frames. Second, the coarse representation is refined by two parallel modules, \ie, Part-level Spatial Modeling and Part-level Temporal Modeling. Part-level Spatial Modeling contains spatial partition and spatial attention, while Part-level Temporal Modeling consists of temporal partition and temporal attention. The two modeling process is complementary. 
	Part-level Spatial Modeling focuses on capturing the appearance-based properties in the spatial space, while Part-level Temporal Modeling is to capture the motion changes from long-short term contexts along the temporal dimension. Finally, we fuse the refined spatial and temporal features with Multilayer Perceptron (MLP) and one linear classification layer for predicting the sign word.
	It is worth noting that we do not require extra key-point annotations throughout the whole training process.
	
	\subsection{Part-level Spatial Modeling}

	Current skeleton-based approaches \cite{lin2022joint} suffer from the inaccuracy of the keypoints, such as motion blur and self-occlusion, and neglect facial expressions. Therefore, we resort to Part-aware RGB-based methods. Unlike most existing RGB-based methods, we explicitly draw the network's attention to local patterns, which can capture the fine-grained positions between two hand gestures, and the subtle facial expression changes.   
	Specifically, we design a module named Part-level Spatial Modeling that can explicitly capture the hand relationships and the hand-face correlation without extra pose estimators. 
	Figure~\ref{framework}~(b) (c) show two components of Part-level Spatial Modeling: \textbf{spatial partition} and \textbf{spatial attention}.
	
	\noindent\textbf{Spatial Partition.} We first introduce the spatial partition strategy in the Part-level Spatial Modeling. Given a sign video $\textbf{X}$ with T frames, $\textbf{X}\in{\mathbb{R}^{T\times C\times H \times W}}$, where $C, H$, and $W$ are the channel, height, and weight of a single frame, we put the frames of the sign video into the backbone, a spatial-temporal extraction model, and then obtain feature maps $\textbf{M}$.
	Here we have two types of operations for feature maps. One is the global guidance, which averagely pools the feature maps $\textbf{M}$ along the channel dimension and outputs the global feature $g_{sg}$ of size $T \times C$.
	The other processing method is called local guidance, which splits $\textbf{M}$ into two stripes and averages the inner vectors of stripes into a single part-level vector. For the left-right partition, the splitting direction
	is vertical, and the part-level vectors are named $h_{l}, h_{r}$. In addition, we deploy a top-bottom partition that horizontally splits $\textbf{M}$, and the part-level vectors are termed as $h_{t}, h_{b}$.
	After that, Considering the different scales of local features, the Gate function is introduced to normalize these part-level features. 
	Finally, we add up these re-scaled features after gating as the intermediate feature $g$. The process can be formulated as follows, 
	
	\begin{equation}
		\begin{aligned}
			g_{lr} = G_{left} (h_{l}) + G_{right} (h_{r}), \\
			g_{tb} = G_{top} (h_{t}) + G_{bottom} (h_{b}),
		\end{aligned}
		\label{gate}
	\end{equation}
	where $	G(h) = h \cdot Sigmoid (MLP (h))$, and $MLP$ denotes Multilayer Perceptron.
	
	\noindent\textbf{Spatial Attention.}
	Spatial attention focus on complementing information capturing the relationship of hands and faces from refined features $g_{lr}, g_{tb}$ to the global feature $g_{sg}$. 
	Specifically, we deploy linear layers for the refined features~$g_{lr}, g_{tb}$ to generate key-value pairs, termed as $K_{lr}, V_{lr}, K_{tb}, V_{tb}$, which encodes the local patterns about hands and faces. Similarly, we generate a query-value pair from the global feature~$g_{sg}$, named $Q_{s}, V_{s}$. Then we concatenate the key and value of the refined features~$g_{lr}, g_{tb}$ as a large key-value pool and further compute the attention maps with the query of the global feature~$g_{sg}$. In this way, we can extract complementary information from the refined features~$g_{lr}, g_{tb}$ by the attention maps. Finally, we add the complementing features to the value of the global feature~$V_{s}$. The operations are defined as follows,
	\begin{equation}
		\begin{aligned}
			f_{s} &= Softmax (Q_{s}K_{p}^T)V_{p} + V_{s},
		\end{aligned}
		\label{ca_sp}	\end{equation}
	where $K_{p}$ is the concatenated feature of $K_{lr}, K_{tb}$.  $V_{p}$ is the concatenated feature of $V_{lr}, V_{tb}$,  and $f_{s}$ is the final representation of the Part-level Spatial Modeling.
	
\noindent\textbf{Discussion.} 
	In this section, we propose a Partition strategy that adaptively mines the location of face and hands by the pooling operation and fuse the normalized part-level features through the Gate mechanism. 
	Instead of cropping hands or face to mine discriminative representation, this method harnesses the spatial properties of pooling that obtain larger receptive fields. It promotes the bottom-top alignment to build the relationship between the hands and face, and left-right alignment to build the relationship between hands. 
	It is also efficient because of no need to preprocess the input, such as the pose estimator. Besides, we also propose an Attention method, which further aggregates the local and global clues. In summary, mutual modeling of the Part-level Spatial Modeling, especially face and hand in an end-to-end way, helps to capture discriminative fine-grained attention.
	
	\subsection{Part-level Temporal Modeling}
	Current SLR methods overlook the bottom-up property that the sign sequences can be viewed as a combination of multiple sub-actions. This property motivates us to explore the short-segment contexts. In Figure~\ref{mo}(b), combining more part sequences as sub-actions can fully exploit the part-level temporal property.
	Hence, we introduce Part-level Temporal Modeling for mining the long-short term action change for sign language. 
	The idea is that we view the sign sequence as a combination of multiple decomposed sub-actions. 
	Specifically, this module learns how to handle small segments to complement the long segment by explicitly segmenting the long segment in the temporal dimension. 
	This module involves two stages: \textbf{temporal partition} and \textbf{temporal attention}.
	
	\noindent\textbf{Temporal Partition.}
	The temporal partition is similar to the spatial partition. The main difference is that we split feature maps along the temporal dimension.
	Given feature maps $\textbf{M}$, we first pool the $(H, W)$ dimension of feature maps to $(1, 1)$ because we mainly model the temporal cues in this branch, which also reduces computing costs.
	Then we segment the temporal dimension $T$ to $N=3$ parts with overlaps as shown in Figure~\ref{framework}~(f), $\textbf{S} = \{s_n\}_{n=1}^{N}$.
	
	\noindent\textbf{Temporal Attention.}
	To mine the temporal motion changes,  we deploy Gated Recurrent Neural Networks ($GRUs$). We apply an independent  $GRU$ on the global feature that captures the long-term change in the whole video. 
	Other GRUs are used for every short segment $s_n$ as follows:
    
	\begin{equation}
		g_n = GRU_{n} (s_n), 
		\label{GRU_tp}
	\end{equation}
	where $GRU_n$ does not share weights, considering the original short sequence orders.
	Next, we calculate the key-value pairs, termed as $K_n, V_n$, through a linear function and concatenate these keys anD values as $K'_p, V'_p$, respectively. Similarly, we compute the query-value pair of the global feature~$g_{t}$, named as $Q_t, V_t$.
	Then we deploy the temporal attention function to calculate the relationship between short segments and the long sequence.
	The attention of this part aims to explore the long-short-term problem. 
	This operation enhances the capacity to model long-short-term variations of hands or faces as follows,
	\begin{equation}
		\begin{aligned}
			f_{t} &= Softmax (Q_t{K'_p}^T)V'_p + V_t.
		\end{aligned}
		\label{ca_tp}
	\end{equation}
	After achieving the spatial and temporal refined features, \ie, $f_{s}, f_{t}$, we apply an MLP on the concatenated feature of $f_{s}$ and $f_{t}$ to derive the final representation $f_{st}$.
	
	\noindent\textbf{Discussion.} 
	In Figure~\ref{mo}, we show that the short video clips alone can facilitate the network to predict the correct sign language class. Inspired by this observation, we explicitly extract short-term representations from short video clips and then fuse multiple short-term predictions as one long-short-term prediction, which facilitates the robust inference process.
	Following the spirit, we propose a simple fusion method based on splitting and aggregation, where the splitting allows our model to fully exploit the fine-grained short-term representation and leverage these features to conduct the attention mechanism.
	The aggregation allows our model to explore the complementary way between the context of short and long segments and handle the long-short range dependencies and make full use of local and global cues.

    \noindent\textbf{Discussion about partition-attention mechanism.} Sign language recognition tasks need to consider both hand movement and facial appearance. Thus mining the fine-grained relations between them both in spatial and temporal dimensions will help the model to learn more discriminate features.
Based on the above observations, we design a spatial partition method that targets to model the local relations between hands and faces separately. After that, the mined fine-grained local feature will complement the global feature by an attention mechanism, yielding a more robust global feature. 
Moreover, simply modeling the spatial relations is not enough to discriminate challenging cases. Thus we further design a temporal modeling branch that summarises the long-short term changes in temporal context, capturing the hands and faces changes. By coupling all the spatial-temporal cues, the model achieves significant improvements.
    
	\subsection{Optimization Objectives}

	Given the class predictions from different spatial and temporal parts, we optimize the classification error following the existing works~\cite{li2020word,jiang2021sign,hu2021hand}, which can be formulated as cross-entropy loss: $\mathcal{L}_{ce} (q, y) = -\log(\frac{exp(q_y)}{\sum_{c=1}^C exp(q_c)}),$ where $q$ is the predicted logits, and $y$ is the index of the ground-truth category. Different from previous works, we accumulate the cross-entropy losses on all partial and global observations, and thus the total loss can be defined as follows,
	\begin{equation}
		\begin{aligned}
			\mathcal{L}_{total} &= 
			\overbrace{ 
				\mathcal{L}_{ce}(q_l,y) +  \mathcal{L}_{ce}(q_r,y) + 
				\mathcal{L}_{ce}(q_t,y) +
				\mathcal{L}_{ce}(q_b,y)
			}^{\mathcal{L}_{spatial}}
			\\
			+& \overbrace{ 
				\mathcal{L}_{ce}(q_{lr},y) +  \mathcal{L}_{ce}(q_{tb},y) + 
				\mathcal{L}_{ce}(q_{sg},y) +
				\mathcal{L}_{ce}(q_{s},y)
			}^{\mathcal{L}_{spatial}} \\
			+& \underbrace{\mathcal{L}_{ce}(q_{t},y)}_{\mathcal{L}_{temporal}} +  \underbrace{\mathcal{L}_{ce}(q_{st},y)}_{\mathcal{L}_{fuse}}
			\label{loss}
		\end{aligned}
	\end{equation}
	where $\mathcal{L}_{spatial}$ is utilized to supervise the predictions from spatial features including $q_l, q_r, q_t, q_b$ as well as the aggregated features, \eg, $q_{lr}, q_{tb}$. Applying loss on $q_l, q_r, q_t, q_b$ can help the model to mine the local discriminate feature so that only seeing one part of these can yield relatively robust results. Moreover, applying loss on the aggregated features $q_{lr}, q_{tb}$ can help the model learn the relations and correlations between the local discriminate feature.
	The classification loss motivates the model to discriminate the video with different sign language meanings, and implicitly encourages the model to learn the symmetric relationship, \eg, between two hands, and the top-down correlation, \eg, between hands and faces. Besides, it also aggregates the features representing global-local human structure by attention mechanism. Similarly, for another temporal branch, we deploy $\mathcal{L}_{temporal}$ on the prediction $q_{t}$ of aggregated feature $f_{t}$ to implicitly capture the long-short term variation. 
	Finally, $\mathcal{L}_{fuse}$ motivates the model to learn adaptive weights between the spatial and the temporal feature $f_{s}$ and $f_{t}$. During inference, we only deploy the top prediction $q^{rgb}_{st}$ based on the final feature as the predicted sign word.
	
	\begin{figure}[!t]
		\centering
		\includegraphics[width=1\linewidth]{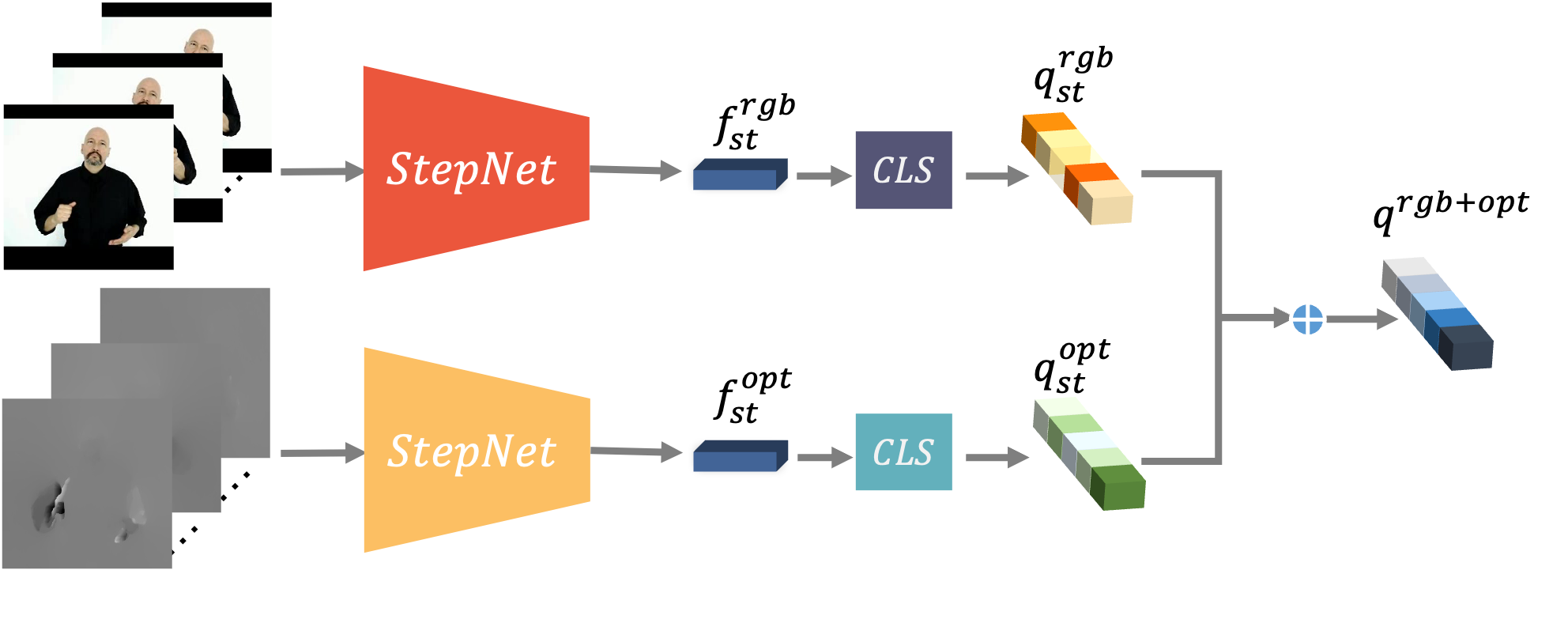}
		\caption{A schematic of our two-stream fusion method. We fuse the classification logits~($q^{rgb}_{st}, q^{opt}_{st}$) processed from two StepNets to obtain the final logits~($q_{r+o}$). \textbf{Notably, two StepNets do not share weights.  
		}
		}
		\label{fusin_net}
	\end{figure}
	
	\subsection{Two-Stream StepNet}
	\label{twostream}
	StepNet is flexible and compatible with inputs of different modalities.
	We can equip multiple StepNets to form a simple late-fusion framework for the multi-modality input. For instance, as shown in Figure~\ref{fusin_net}, we introduce another modality, \ie, optical flow, which captures the motion between frames. We separately train the two StepNets for RGB and optical flow, of which weights are not shared. 
	We adopt the late-fusion strategy to combine the category predictions. 
	Specifically, we sum up the two modality logits~($q^{rgb}_{st}, q^{opt}_{st}$), generated from two StepNets with weights to obtain the final prediction: 
	\begin{equation}
		q^{rgb+opt} = q^{rgb}_{st} + \alpha q^{opt}_{st},
		\label{fusion_func}
	\end{equation}
	where $q$ denotes the prediction logits, $\alpha$ is a weight hyper-parameter. For instance, we empirically set $\alpha = 0.4$ for the WLASL. Albeit simple, we show that the late-fusion strategy can arrive at a competitive sign language recognition accuracy in the experiment. 
	
\section{Experiments} 
	\begin{table}[!t]
		\caption{A statistical summary of SLR datasets.}
		\label{dataset_summary}
		\tabcolsep=2pt
		\centering
		\begin{tabular}{l | c c r c c}
			\toprule[2pt]
			Datasets &  \#  Signs &  \#  Signers & \#  Samples & Languages & Type \\
			\midrule[1pt]
			WLASL~\cite{li2020word}   & 2,000 & 119 & 21K  & American & Isolated \\
			NMFs-CSL~\cite{hu2021global}   & 1,067 & 10  & 32K & Chinese & Isolated\\
			BOBSL~\cite{albanie2021bbc}    & 2,281 & 39  & 452K & British & Co-articulated\\
			\bottomrule[1pt]
		\end{tabular}
	\end{table}
	
	\subsection{Datasets}
	We evaluate the proposed StepNet on three widely-used public datasets, including WLASL~\cite{li2020word}, NMFs-CSL~\cite{hu2021global}, BOBSL~\cite{albanie2021bbc}, which span different sign languages. The datasets are briefly summarized in TABLE \ref{dataset_summary}.
	
	\noindent\textbf{WLASL~\cite{li2020word}} is an American Sign Language~(ASL) dataset, which includes four subsets consisting of different vocabulary sizes, \ie, WLASL-100, WLASL-300, WLASL-1000, WLASL-2000. For instance, WLASL contains 21,083 videos performed by 119 signers and originated from unrestricted videos on the web. 
	WLASL-2000 is one of the most challenging subsets because of more infrequent words. 
	
	\noindent\textbf{NMFs-CSL~\cite{hu2021global}} is a Chinese Sign Language~(CSL) dataset, which includes 25,608 and 6,402 samples with 1,067 words for training and testing, respectively. It is also a challenging dataset due to a large variety of confusing words.
	
	\noindent\textbf{BOBSL~\cite{albanie2021bbc}} is a large-scale video collection of British Sign Language~(BSL), which contains 1,962 episodes spanning a total of 426 differently named TV shows. It has 452K samples performed by 39 different signers for Co-articulated sign language recognition. This dataset differs from the above datasets because it is under the Co-articulated signing setting, which signs in context. It benefits to build robust models for understanding sign language ``in the wild''.
	
	Note that all these datasets employ the signer-independent setting, which means no signer overlap between the training and testing splits.

	\subsection{Implementation Details}
	We apply FFmpeg~\cite{tomar2006converting} to extract all frames from RGB videos and resize all frames to 320$\times$256. 
	During training, we uniformly split the video into 16 splits and then randomly sample 16 frames from each split.
    The frames are then randomly cropped to 256$\times$256. When testing, we center sample 16 frames and then center crop the frames to 256$\times$256.
	We also apply the random horizontal flip augmentation to frames with a probability of 0.5. 
	StepNet is implemented by PyTorch~\cite{paszke2017automatic} and trained on one NVIDIA TESLA V100 with a batch size of 8. We train our framework using the AdamW~\cite{loshchilov2017decoupled} optimizer. For WLASL~\cite{li2020word} and NMFs-CSL~\cite{hu2021global}, the weight decay is set to 0.1 and the epoch is set to 100. 
	In the training phase, we use linear warm-up in the first five epochs. 
	The initial learning rate is 1e-4, and we reduce it to 1e-5 by CosineAnnealing~\cite{loshchilov2017sgdr} learning rate scheduler. 
	For BOBSL~\cite{albanie2020bsl}, we follow the existing work~\cite{albanie2020bsl}, and deploy SGD optimizer with 0.9 momentum. The learning rate and the epoch are set to 0.03 and 30. We apply a multistep scheduler to decay the learning rate. 
	The decay parameter $\gamma$ and step are set to 0.1 and [15, 25], respectively.
	For multi-modality fusion, we follow~\cite{lin2019tsm} to extract optical flow data, and adopt the TVL1 algorithm~\cite{zach2007duality} implemented by Openmmlab Denseflow API~\cite{denseflow}. 
The model parameter is 194M. We use TSM (R50) as our backbone, and the spatial and temporal attentions are composed of several linear layers, activation layers, and normalization layers. The running time is 0.0346s per batch (8 bach size) on Tesla V100 (16G). The shape of the tensor is listed as shown in Table \ref{shape_tensor}.
\begin{table*}
		\small
		\tabcolsep=4 pt
            \caption{
			Comparison with the state-of-the-art methods in terms of Top-1, Top-2, and Top-5 accuracy on the NMFs-CSL ~\cite{hu2021global}. The best results are in \textbf{bold}. 
		\emph{H} denotes the reconstructed hand representation by MANO~\cite{MANO:SIGGRAPHASIA:2017}.
		\emph{H}+\emph{P} and \emph{H}+\emph{R} mean fusing the reconstructed hand representation (\emph{H}) to the features of the Skeleton-based~(\emph{P}) and the RGB-based~(\emph{R}) method, respectively. \emph{R}+\emph{F} denotes using RGB data and optical flow data as inputs.}
            \resizebox{\textwidth}{!}{
			\begin{tabular}{l|ccc|ccc|ccc}
				\toprule[2pt]
				\multirow{2}{*}{Method}      &  \multicolumn{3}{c|}{Total} &  \multicolumn{3}{c|}{Confusing} &  \multicolumn{3}{c}{Normal}  \\ 
				& Top-1 & Top-2 & Top-5 & Top-1 & Top-2 & Top-5 & Top-1 & Top-2 & Top-5 \\ 
				\midrule[1pt]
				\midrule[1pt]
				\textbf{Skeleton-based}  & & & & & & & & & \\
				ST-GCN~\cite{yan2018spatial} & 59.9 & 74.7 & 86.8 & 42.2 & 62.3 & 79.4 & 83.4 & 91.3 & 96.7 \\
				Signbert~(\emph{H})~\cite{hu2021signbert}  & 67.0 & 86.8 & 95.3 & 46.4 & 78.2 & 92.1 & 94.5 & 98.1 & 99.6 \\
    BEST~\cite{zhao2023best}   & 68.5 &- &94.4  &49.0 &- &90.3 &94.6 &- &99.7 \\
				\midrule[1pt]
				
				\textbf{RGB-based}  & & & & & & & & & \\
				3D-R50~\cite{qiu2017learning}  & 62.1 & 73.2 & 82.9 & 43.1 & 57.9 & 72.4 & 87.4 & 93.4 & 97.0 \\
				DNF~\cite{cui2019deep}         & 55.8 & 69.5 & 82.4 & 33.1 & 51.9 & 71.4 & 86.3 & 93.1 & 97.0 \\
				I3D~\cite{carreira2017quo}     & 64.4 & 77.9 & 88.0 & 47.3 & 65.7 & 81.8 & 87.1 & 94.3 & 97.3 \\
				TSM~\cite{lin2019tsm}          & 64.5 & 79.5 & 88.7 & 42.9 & 66.0 & 81.0 & 93.3 & 97.5 & 99.0 \\
				Slowfast~\cite{feichtenhofer2019slowfast} & 66.3 & 77.8 & 86.6 & 47.0 & 63.7 & 77.4 & 92.0 & 96.7 & 98.9 \\ 
				GLE-Net~\cite{hu2021global}  & 69.0 & 79.9  & 88.1 & 50.6 & 66.7 & 79.6 & 93.6 & 97.6 & 99.3 \\ 
				Ours & \textbf{77.2} \textcolor{green}{$\uparrow$ 8.2} & \textbf{86.2} \textcolor{green}{$\uparrow$ 6.3}  & \textbf{92.5} \textcolor{green}{$\uparrow$ 3.8} & \textbf{62.4} \textcolor{green}{$\uparrow$ 11.8} & \textbf{76.1} \textcolor{green}{$\uparrow$ 9.4} & \textbf{86.9} \textcolor{green}{$\uparrow$ 5.9} & \textbf{96.9} \textcolor{green}{$\uparrow$ 3.3} & \textbf{99.7}  \textcolor{green}{$\uparrow$ 2.1} & \textbf{99.9}  \textcolor{green}{$\uparrow$ 0.6} \\
				\midrule[1pt]
				
				\textbf{Fusion-based}  & & & & & & & & & \\
				Signbert~ (\emph{H} + \emph{P})~\cite{hu2021signbert} & 74.9 & \textbf{93.2} & \textbf{98.2} & 58.6 & \textbf{88.6} & \textbf{96.9} & 96.7 & 99.3 & 99.9 \\ 
				Signbert~ (\emph{H} + \emph{R})~\cite{hu2021signbert}& 78.4 & 92.0 & 97.3 & 64.3 & 86.5 & 95.4 & 97.4 & 99.3 & 99.9 \\ 
                BEST (\emph{H} + \emph{R})~\cite{zhao2023best}  &79.2 &- &97.1 &65.5 &- &95.0 &97.5 &- &99.9 \\
				Ours~ (\emph{R}+\emph{F})& \textbf{83.6} \textcolor{green}{$\uparrow$ 4.4}  & 92.7 \textcolor{red}{$\downarrow$ 0.5} & 97.0 \textcolor{red}{$\downarrow$ 1.2} & \textbf{72.3} \textcolor{green}{$\uparrow$ 6.8} & 87.2 \textcolor{red}{$\downarrow$ 1.4} & 94.8 \textcolor{red}{$\downarrow$ 2.1} & \textbf{98.7} \textcolor{green}{$\uparrow$ 1.2} & \textbf{99.9} \textcolor{green}{$\uparrow$ 0.6} & \textbf{100.0} \color{green}{$\uparrow$ 0.1} \\ 
				\bottomrule[1pt]
			\end{tabular}
            }
            
		\label{NMFs-CSL}
	\end{table*}
	
    \begin{table}[h]
    \centering
    \small
    \tabcolsep=4pt
    \caption{
    The shape of tensors.
    }
        \begin{tabular}{l | c | c | c|c|c|c|c|c|c|c|c|c|c|c|c }
            \toprule[2pt]
             name & $M$ & $f_s$ &  $f_t$ & $f_{st}$ & $h_l$ & $h_r$ & $h_t$ & $h_b$ & $g_{lr}$ & $g_{tb}$ & $g_{sg}$ & $g_{1}$ &  $g_{2}$ & $g_{3}$ & $g_{t}$\\
            \midrule[1pt]
             size &  16 $\times$ 2048 $\times$ 16 $\times$ 16 &
              \multicolumn{2}{c|}{16 $\times$ 1024} &
              \multicolumn{8}{c|}{16 $\times$ 2048} &
            \multicolumn{3}{c|}{8 $\times$ 1024} &
                 16 $\times$ 2048  \\
                
            \bottomrule[1pt]
        \end{tabular}
        \label{shape_tensor}
    \end{table}
	
    \begin{table}
        \small
        \tabcolsep=1 pt
        \caption{
        Comparison with the state-of-the-art methods in terms of Top-1 and Top-5 accuracy on the WLASL~\cite{li2020word}. The best results are in \textbf{bold}. Per-i denotes Per-instance, and Per-c represents Per-class. \emph{H} denotes the reconstructed hand representation by MANO~\cite{MANO:SIGGRAPHASIA:2017}. \emph{P} means using skeleton data as input.
        }
        \resizebox{\textwidth}{!}{
            \begin{tabular}{l|cc|cc|cc|cc|cc|cc|cc|cc}
            \toprule[2pt]
            \multirow{3}{*}{Method} 
            & \multicolumn{4}{c|}{WLASL100~\cite{li2020word}}
            & \multicolumn{4}{c|}{WLASL300~\cite{li2020word}}
            & \multicolumn{4}{c|}{WLASL1000~\cite{li2020word}}
            & \multicolumn{4}{c}{WLASL2000~\cite{li2020word}}\\ \cline{2-17}
            & \multicolumn{2}{c|}{Per-i} & \multicolumn{2}{c|}{Per-c} 
            & \multicolumn{2}{c|}{Per-i} & \multicolumn{2}{c|}{Per-c}
            & \multicolumn{2}{c|}{Per-i}   & \multicolumn{2}{c}{Per-c}
            & \multicolumn{2}{c|}{Per-i} & \multicolumn{2}{c}{Per-c}
            \\
            & Top-1 & Top-5 & Top-1 & Top-5 
            & Top-1 & Top-5 & Top-1 & Top-5   
            & Top-1 & Top-5   & Top-1 & Top-5 
            & Top-1 & Top-5 & Top-1 & Top-5 \\ \midrule[1pt] \midrule[1pt]
            \textbf{Skeleton-based} & & & & & & & & & & & & & & \\
            ST-GCN~\cite{yan2018spatial} & 50.78 & 79.07 & 51.62 & 79.47   
            & 44.46 & 73.05 & 45.29 & 73.16
            & - & - & - & -
            & 34.40 & 66.57 & 32.53 & 65.45 \\ 
            Pose-TGCN~\cite{li2020word} & 55.43 & 78.68 & - & -   
            & 38.32 & 67.51 & - & -
            & - & - & - & -
            & 23.65 & 51.75 & - & - \\ 
            PSLR~\cite{tunga2020pose} & 60.15 & 83.98 & - & -   
            & 42.18 & 71.71 & - & -
            & - & - & - & -
            & - & - & - & - \\ 
            Signbert~(\emph{H})~\cite{hu2021signbert} & 76.36 & 91.09 & 77.68 & 91.67  
            & 62.72 & 85.18 & 63.43 & 85.71 
            & - & - & - & -
            & 39.40 & 73.35 & 36.74 & 72.38 \\ 
                BEST~\cite{zhao2023best} &  77.91 & 91.47 &77.83 &92.50 &67.66 &89.22 &68.31 &89.57& - & - & - & -&46.25 &79.33 &43.52 &77.65 \\
            Signbert+~\cite{Hu_2023} & 79.84 & 92.64 & 80.72 & 93.08 &73.20 &90.42 &73.77 &90.58 &- &-& - & - &48.85 &82.48 &46.37 &81.33 \\
            SAM~(\emph{P})~\cite{jiang2021sign}      & - & - & - & -  
            & - & - & - & - 
            & - & - & - & -
            & 51.50 & 84.94 & 48.87 & 84.02 \\ \midrule[1pt]
            
            \textbf{RGB-based} & & & & & & & & & & & & & &\\
                MEN~\cite{9781811}  & - & - & - & -  
            & - & - & - & -
            & - &- & - & -
            & 25.54 & 53.72 & - & - \\
            I3D~\cite{li2020word}  & 65.89 & 84.11 & 67.01 & 84.58  
            & 56.14 & 79.94 & 56.24 & 78.38
            & 47.33 &76.44 & - & -
            & 32.48 & 57.31 & - & - \\ 
            TCK~\cite{li2020transfer} & 77.52 & 91.08 & 77.55 & 91.42   
            & 68.56 & 89.52 & 68.75 & 89.41
            & - & - & - & -
            & - & - & - & - \\ 
            BSL~\cite{albanie2020bsl}  & - & - & - & -  
            & - & - & - & -
            & - & -& - & -
            & 46.82 & 79.36 & 44.72 & 78.47 \\ 
            
                Ours   & \textbf{78.29} \textcolor{green}{$\uparrow$ 0.8} & \textbf{92.25} \textcolor{green}{$\uparrow$ 1.2}
                & \textbf{78.77} \textcolor{green}{$\uparrow$ 1.2} & \textbf{92.63} \textcolor{green}{$\uparrow$ 1.2}
                & \textbf{74.70} \textcolor{green}{$\uparrow$ 6.1} & \textbf{91.02} \textcolor{green}{$\uparrow$ 1.5}  & \textbf{75.32} \textcolor{green}{$\uparrow$ 6.6} & \textbf{91.17} \textcolor{green}{$\uparrow$ 1.8} 
                & \textbf{67.91} & \textbf{91.10} & \textbf{67.76} & \textbf{91.33}&
                \textbf{56.89} \textcolor{green}{$\uparrow$ 10.1} & \textbf{88.64} \textcolor{green}{$\uparrow$ 9.3} & \textbf{54.54} \textcolor{green}{$\uparrow$ 9.8} & \textbf{87.97} \textcolor{green}{$\uparrow$ 9.5} \\ 
            \bottomrule[1pt]
        \end{tabular}
        }

        \label{wlasl}
    \end{table}
 
	\subsection{Quantitative Results}
	To validate the effectiveness of our model, we use Per-instance and Per-class accuracy metrics, which mean the average accuracy over each instance and each class, respectively. 
	For NMFs-CSL~\cite{hu2021global}, we follow~\cite{hu2021global} and report the Per-instance accuracy including Top-1, Top-2, Top-5 accuracy. For WLASL~\cite{li2020word} and BOBSL~\cite{albanie2021bbc}, we show Per-instance and Per-class metrics with the Top-1 and Top-5 accuracy. 
	
	\noindent \textbf{Comparison with state-of-the-art methods.}
	We compare the proposed method with several competitive SLR methods on three benchmark datasets, \ie, NMFs-CSL~\cite{hu2021global}, WLASL~\cite{li2020word} and BOBSL~\cite{albanie2021bbc}.
	
	\noindent \textbf{Evaluation on NMFs-CSL~\cite{hu2021global}.} 
	As shown in Table~\ref{NMFs-CSL}, we obtain the best accuracy on total, normal and confusing words, which validates the effectiveness of our proposed method. Specifically, when only one modality is employed, our RGB-based StepNet surpasses the best Skeletion-based method, \ie, Signbert~\cite{hu2021signbert}, by a large margin, \ie, +10.2\% Top-1 increment. Meanwhile, compared with the best RGB-based GLE-Net~\cite{hu2021global}, our method acquires an +8.2\% improvement in Top-1 accuracy. For comparison with the fusion-based method, we implement the two-stream framework using RGB and optical flow data, termed as \emph{R}+\emph{F} network (See Section \ref{twostream} for more details).
	We also achieve very competitive performance compared with the fusion-based method.
	
	\noindent \textbf{Evaluation on WLASL~\cite{li2020word}.} WLASL has four subsets, \ie, WLASL-100, WLASL-300, WLASL-1000, and WLASL-2000. The number in the subset represents the number of words provided within the subset.
	TABLE~\ref{wlasl} shows that our proposed method outperforms the previous state-of-the-art method on all subsets by a clear margin.  
	In particular, our method surpasses the recent Skeleton-based SAM~\cite{jiang2021sign} with +5.39\% Top-1 Per-instance improvement and +5.67\% Top-1 Per-class improvement. 
	We also implement StepNet (Optical flow) mentioned above.
	As shown in TABLE~\ref{WLASL_fusion}, when fusing with the optical flow network, we achieve 61.17\% Top-1 er-instance accuracy and 91.94\% Top-5 Per-instance accuracy, which outperform the SAM~\cite{jiang2021sign} (7 modalities + 5 clips). It is worth noting that we only use two modalities (RGB + Optical flow) to improve our model and only use one clip during inference. Our proposed method surpasses the previous state-of-the-art method with fewer modality inputs. 
	\begin{table}[!t]
		\setlength{\tabcolsep}{25 pt}
		\caption{Comparison with the state-of-the-art fusion-based methods in terms of Top-1 and Top-5 accuracy on WLASL-2000~\cite{li2020word}. The best results are in \textbf{bold}. \emph{C} denotes the five-clip ensemble method. \emph{H} is the reconstructed hand representation by MANO~\cite{MANO:SIGGRAPHASIA:2017}. \emph{H}+\emph{P} or \emph{H}+\emph{R} in the Signbert mean fusing the reconstructed hand representation (\emph{H}) to the features of the Skeleton-based or RGB-based method. \emph{R}+\emph{F} denotes using RGB data and optical flow data as inputs. }
		\label{WLASL_fusion}
		\begin{center}
                \resizebox{\textwidth}{!}{
			\begin{tabular}{l | cc | cc }
				
				\toprule[2pt]
				\multirow{2}{*}{Model} & \multicolumn{2}{c|}{Per-instance} & \multicolumn{2}{c}{Per-class} \\ 
				& Top-1 & Top-5 & Top-1 & Top-5 \\
				\midrule[1pt]
				Signbert (\emph{H}+\emph{P})~\cite{hu2021signbert} 
				& 47.46 & 83.32 & 45.17 & 82.32 \\ 
				Signbert (\emph{H}+\emph{R})~\cite{hu2021signbert}  
				& 54.69 & 87.49 & 52.08 & 86.93 \\ 
            BEST~\cite{zhao2023best} &54.59 & 88.08 &52.12 &87.28 \\
                   Signbert+~\cite{Hu_2023} & 55.59 &89.37 &53.33 &88.82 \\
				SAM (7 Modalities + \emph{C})~\cite{jiang2021sign} 
				& 59.39 & 91.48 & 56.63 & 90.89 \\ 
				Ours (\emph{R}+\emph{F}) 
            & \textbf{61.17} \textcolor{green}{$\uparrow$ 1.78} & \textbf{91.94} \textcolor{green}{$\uparrow$ 0.46} & \textbf{58.43} \textcolor{green}{$\uparrow$ 1.8} & \textbf{91.43} \textcolor{green}{$\uparrow$ 0.54} \\ 
				\bottomrule[1pt]
			\end{tabular}
            }
		\end{center}
		
	\end{table}
	
	\begin{table}[!ht]
		\setlength{\tabcolsep}{28 pt}
		\caption{Comparison with the state-of-the-art methods in terms of Top-1 and Top-5 accuracy on the BOBSL~\cite{albanie2021bbc}. The best results are in \textbf{bold}.}
		\label{bobsl}
		\begin{center}
                \resizebox{\textwidth}{!}{
			\begin{tabular}{l | cc | cc }
				\toprule[2pt]
				\multirow{2}{*}{Model} & \multicolumn{2}{c|}{Per-instance} & \multicolumn{2}{c}{Per-class} \\ 
				& Top-1 & Top-5 & Top-1 & Top-5 \\
				\midrule[1pt]
				2D pose-Sign~\cite{albanie2021bbc} &61.8 &82.1 &30.6 &56.6 \\
				Flow-I3D~\cite{albanie2021bbc} &52.1 &75.7 &19.2 &41.7 \\
				RGB-I3D~\cite{albanie2021bbc} &75.8 &92.4 &50.5 &77.6 \\
				Ours~(I3D) & \textbf{77.1} & \textbf{92.7} & \textbf{51.3} & \textbf{78.2}\\
				\bottomrule[1pt]
			\end{tabular}
                }
		\end{center}
		
	\end{table}

	\noindent \textbf{Evaluation on BOBSL~\cite{albanie2021bbc}.} 
	BOBSL is a new dataset containing more words and sign videos than WLASL-2000~\cite{li2020word}. We observe similar performance improvement.
	As shown in Table~\ref{bobsl}, our implemented StepNet~(I3D) achieves 77.1\% Top-1 accuracy and 92.7\% Top-5 accuracy in Per-instance metrics, which boosts the baseline(I3D) by 1.3\% Top-1 and 0.3\% Top-5 accuracy. In Per-class metrics, we also obtain the improvement in Top-1 and Top-5 accuracy. Specifically, Our network arrives at 51.3\% Top-1 and 78.2\% Top-5 accuracy. 

	\subsection{Ablation Studies}
	To validate the effectiveness of the proposed method, we design the following ablation studies, \ie, spatial and temporal clues, changing backbones and the fusion ratio. If not specified, we mainly evaluate the model on the widely-used WLASL-2000.
	
	\noindent\textbf{Effect of spatial and temporal clues.} As shown in Table~\ref{ablation on stb}, we have several observations: 1) our model based on TSM~\cite{lin2019tsm} obtains +1.88\% Top-1 accuracy improvement when we merely use the proposed spatial clues.
    2) when only using temporal cues, our model also obtains 1.04\% Top-1 improvement compared with the baseline (TSM).
	3) when fusing spatial and temporal clues, our model obtains 56.89\% Top-1 per-instance accuracy, which brings 2.88\% accuracy improvement. It also shows that spatial and temporal cues are complementary.

	\begin{table}[!t]
		\caption{Ablation studies on WLASL-2000~\cite{li2020word}.}
            \begin{subtable}[t]{0.45\textwidth}
            \subcaption{
            Two proposed branches. $w$ denotes that we harness with such a component.
            }
            \label{ablation on stb}
            \setlength{\tabcolsep}{2 pt}
            \resizebox{\textwidth}{!}{
            \begin{tabular}{l | c | c | c}
                \toprule[2pt]
                Backbone & $w/~ Spatial$ & $w/~ Temporal$ &  Top-1\\
                \midrule[1pt]
                \multirow{5}{*}{I3D + Ours} & & & 43.38\\
                & \cmark & &  46.06 (+2.68)\\
                & & \cmark & 48.52 (+5.14) \\
                & \cmark & \cmark & 49.46 (+6.08) \\
                \midrule[1pt]
                \multirow{4}{*}{TSM + Ours} & & & 54.01\\
                & \cmark & & 55.89  (+1.88)\\
                & & \cmark & 55.05 (+1.04)\\
                & \cmark & \cmark & 56.89 (+2.88)\\
                \bottomrule[1pt]
            \end{tabular}
            }
            \end{subtable}
            \hfill
            \begin{subtable}[t]{0.45\textwidth}
            \subcaption{Part-level Spatial Modeling. $tb$ means top-bottom partition, while $lr$ denotes left-right partition.}
            \label{ab_sp}
            \setlength{\tabcolsep}{5 pt}
            \resizebox{\textwidth}{!}{
			\begin{tabular}{ l | c | c | c | c}
				\toprule[2pt]
				$tb$ & $lr$ & $attention$ & $concatenate$ & Top-1\\
				\midrule[1pt]
				& & & & 54.01\\
				\cmark & & & & 54.15 (+0.14) \\
				& \cmark & & &  54.19 (+0.18)\\
				\cmark & \cmark & & \cmark &  55.05 (+1.04)\\
				\cmark & \cmark & \cmark & & 55.89 (+1.88) \\
				\bottomrule[1pt]
			\end{tabular}
            }
		\end{subtable}
	\end{table}

	\noindent\textbf{Different backbones.} To validate the scalability of our proposed Part-level Spatial and Temporal modeling methods, we explore the influence of the different backbones. We re-implement the I3D backbone for adapting our proposed method and report the re-implemented I3D performance. 
	As shown in Table~\ref{ablation on stb}, our method also yields about +6.08\% accuracy improvement on the I3D backbone. 
 
	\noindent\textbf{Effect of Spatial and Temporal submodules.}
	In the Part-level Spatial Modeling, we ablate the spatial partition and attention. 1) As shown in Table~\ref{ab_sp}, we study the left-right and top-bottom partition strategies. The result shows that left-right and top-bottom clues improve the baseline by 0.14\% and 0.18\%, respectively. 
	2) To validate the effectiveness of spatial attention, we implement a simple concatenation method for fusing the left-right and top-bottom features. 
	We observe that the attention method surpasses the concatenate method by 0.84\% accuracy in Table~\ref{ab_sp}. It shows that the attention method builds the relationship between the global and local clues.
	3) As shown in Table~\ref{ab_tp}, applying any number of segments or length of segments can improve the baseline (54.01\% Top-1 accuracy). 4) Four splits generally obtain better performance, and setting the appropriate length of sub-clips, such as 6, also can lead to better accuracy.
 
	\noindent\textbf{Effect of GRUs.} 
	As shown in Table~\ref{ab_gru}, applying GRUs on temporal split features improves the performance before using temporal attention. We consider that this operation helps the model to learn the temporal information that represents the sub-action and global-action of signers.
 
\begin{figure}[h]
    \begin{minipage}[!t]{0.42\textwidth}
        \centering
        \begin{tabular}{l | c | c}
            \toprule[2pt]
            StepNet (TSM~\cite{lin2019tsm}) & Top-1 & Top-5\\
            \midrule[1pt]
            $cls_{t}$ & 48.45 &82.18 \\
            $cls_{b}$ & 49.84 &83.88 \\
            $cls_{tb}$ & 54.01 &87.95 \\
            $cls_{l}$ & 51.23 &85.59 \\
            $cls_{r}$ & 48.73 &83.81 \\
            $cls_{lr}$ & 53.94 &87.95  \\
            $cls_{sg}$ & 54.22 &87.81 \\
            $cls_{s}$ & 55.96 &88.64  \\
            $cls_{t}$ & 56.69 &88.57 \\
            $cls_{st}$ & 56.89 &88.64 \\
            \bottomrule[1pt]
        \end{tabular}
        \caption{The Per-instance accuracy of each classifier on WLASL-2000~\cite{li2020word}.}
        \label{cls-stat}
    \end{minipage}%
    \hspace{0.1\textwidth}
    \begin{minipage}[!t]{0.42\textwidth}
        \centering
        \includegraphics[width=1.0\linewidth]{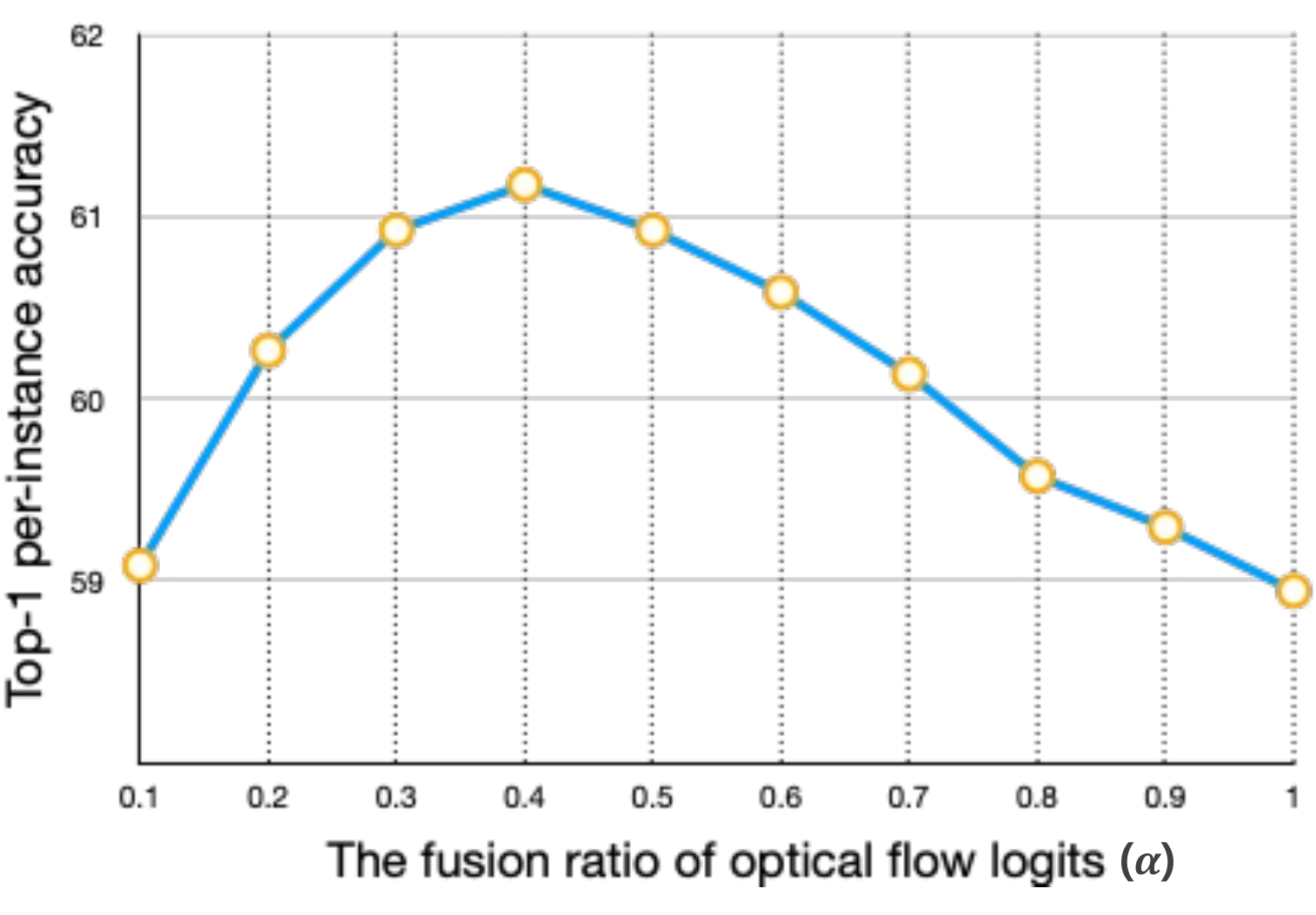}
		\caption{
			The performance of our StepNet~(\emph{R}+\emph{F}) when fusing different ratios on WLASL-2000~\cite{li2020word}. The horizontal axis donates the ratio $\alpha$ in Eq.\eqref{fusion_func}.
		}
		\label{fusion_ratio}
    \end{minipage}
\end{figure}

	\noindent\textbf{Effect of different parts.} In TABLE \ref{cls-stat}, we observe 1) The classifier on local parts also achieves a competitive performance. 2) The global representation, based on local parts, usually achieves a consistent improvement. For instance, $cls_{lr}$ is better than either $cls_l$ or $cls_r$.  
	3) The final classifier on the fused spatial and temporal representation has arrived at the best Top-1 and Top-5 accuracy. Moreover, we conduct experiments on the local temporal features. Our model achieves 56.93 in top-1, and 87.98 in top-5 accuracy. We can see that introducing supervision of short-term temporal features is also helpful to the model.

		
		
	
	\begin{table}[!t]
		
		\caption{Ablation studies of our proposed Part-level Temporal Modeling.}
		
		\begin{center}
                \begin{subtable}[t]{0.5\textwidth}
                \subcaption{Lengths of the segmented frames and segments.}
                \setlength{\tabcolsep}{5 pt}
                \resizebox{\textwidth}{!}{
			\begin{tabular}{l | c | c | c}
				\toprule[2pt]
				Temporal & 2 Segments & 3 Segments & 4 Segments \\
				\midrule[1pt]
				4 Frames & - & - & 55.20 (+1.19) \\
				6 Frames & - & 55.19 (+1.18) & \textbf{55.68 (+1.67)}\\
				8 Frames & 55.30 (+1.29) & 55.08 (+1.07) & 55.33 (+1.32) \\
				\bottomrule[1pt]
			\end{tabular}}
                \label{ab_tp}
                \end{subtable}
                \hfill
                \begin{subtable}[t]{0.4\textwidth}
                \subcaption{Effect of GRUs.}
                \setlength{\tabcolsep}{17 pt}
                \resizebox{\textwidth}{!}{
			\begin{tabular}{ c | c | c}
				\toprule[2pt]
				Type & Top-1& Top-5\\
				\midrule[1pt]
				    w/o GRUs & 55.23 &	87.08\\
				w/ GRUs & \textbf{56.89} & \textbf{88.66}\\
				\bottomrule[1pt]
			\end{tabular}}
                \label{ab_gru}
                \end{subtable}
		\end{center}
	\end{table}
 
	\noindent\textbf{Effect of fusing the optical flow.} 
	Optical flow is able to complement the motion changes between frames for RGB data.
	To compare with the other fusion-based method, we follow~\cite{lin2019tsm} and introduce a StepNet trained on optical flow, which simply replaces the 3-channel RGB inputs with the 10-channel optical flow as discussed in Section~\ref{twostream}. StepNet (Optical Flow) alone reaches a competitive performance 51.2\% Top-1 per-instance accuracy on WLASL-2000~\cite{li2020word}.
	We apply the late-fusion strategy to fuse predictions of StepNet~(RGB) and StepNet~(Optical flow).
	As shown in Figure~\ref{fusion_ratio}, the final prediction is not sensitive to the fusion rate. 
	Fusing optical flow 
	can always improve our RGB network (56.89\%). Therefore, when applying to an unseen situation, RGB and Optical StepNet with a 1:0.3 to 1:0.5 ratio can be a good initial option.
 

\section{Conclusion} \label{conclusion}
	We identify the challenge of underexplored spatial and temporal clues in sign language recognition. 
	As an attempt to fill the gap, we propose a new framework, named as StepNet, by constructing two branches, \ie, Part-level Spatial Modeling and Part-level Temporal Modeling.
	The Part-level Spatial Modeling learns the symmetric association, \eg, between hands and the top-bottom relationship, \eg, between the hands and faces, while the Part-level Temporal Modeling implicitly captures the long-short-term changes. 
	As a result, we achieve competitive performance on three large-scale WLASL, BOBSL, and NMFs-CSL benchmarks, which verifies the effectiveness of the proposed method. 
	In the future, we will continue to study the potential of the proposed method in other fields, such as 3D person re-identification~\cite{zheng2022person}, cooking videos with instruction~\cite{Miech_2019_ICCV}, first-view action recognition~\cite{lu2019deep}, and video question answering~\cite{xue2017unifying, liu2022instance}. 
 
\noindent\textbf{Limitations.} Some of the used networks are out of date, thus the performance may not be better than based on new technology, like transformers. 
Although the model can handle the optical flow data, the performance may not be better than others designed for optical flow. 

\noindent\textbf{Future Enhancements.} Recently, transformers~\cite{li2023catr,shen2023global,ma2023vista} have dominated large numbers of domains of computer vision. Thus designing a transformer-based architecture may enhance the vision representation and further improve the performance. Understanding sign language using LLMs~\cite{yang2024doraemongpt, liu2023llava} may be a new research area. Moreover, designing a pretraining and finetuning schema is also a good solution like BEST~\cite{zhao2023best}, Signbert~\cite{hu2021signbert}.
Our current fusing strategy is based on late fusion which is relatively simple. Designing an effective fusing module is an enhancement for sign language recognition. 

{
\footnotesize
\bibliographystyle{ACM-Reference-Format}
\bibliography{egbib}
}

\end{document}